\documentclass[letterpaper, 10 pt, conference]{ieeeconf}  % Comment this line out if you need a4paper

\IEEEoverridecommandlockouts                              % This command is only needed if 
\usepackage{graphics} % for pdf, bitmapped graphics files
\usepackage{epsfig} % for postscript graphics files
\usepackage{mathptmx} % assumes new font selection scheme installed
\usepackage{times} % assumes new font selection scheme installed
\usepackage{amsmath} % assumes amsmath package installed
\usepackage{amssymb}  % assumes amsmath package installed
\usepackage[mathscr]{euscript}
\usepackage{array}
\usepackage{multirow}
\usepackage{booktabs}
\usepackage{siunitx}
\usepackage{etoolbox} 
\usepackage{tikz}

\newcommand\copyrightnotice[1]{
    \begin{tikzpicture}[remember picture,overlay]
    \node[anchor=south,yshift=10pt] at (current page.south) {\fbox{\parbox{\dimexpr\textwidth-\fboxsep-\fboxrule\relax}{#1}}};
    \end{tikzpicture}
}

\title{\LARGE \bf
Evaluation of Non-collocated Force Feedback \\ 
Driven by Signal-independent Noise
}

\author{Zonghe Chua, Allison M. Okamura, and Darrel R. Deo
\thanks{
Z. Chua, A. M. Okamura, and D. R. Deo are with the Department of Mechanical Engineering, Stanford University, Stanford, CA, 94305. E-mail: ddeo@stanford.edu, chuazh@stanford.edu, and aokamura@stanford.edu \newline \indent *\;This work was supported in part by a Stanford Bio-X fellowship.}% 
}

\begin{document}

% preprint copyright notice
\maketitle

\copyrightnotice{\scriptsize \copyright \, 2020 IEEE. Personal use of this material is permitted. Permission from IEEE must be obtained for all other uses, in any current or future media, including reprinting/republishing this material for advertising or promotional purposes,creating new collective works, for resale or redistribution to servers or lists, or reuse of any copyrighted component of this work in other works.}
%%%%%%%%%%%%%%%%%%%%%%%%%%%%%%%%%%%%%%%%%%%%%%%%%%%%%%%%%%%%%%%%%%%%%%%%%%%%%%%%
\begin{abstract}

Individuals living with paralysis or amputation can operate robotic prostheses using input signals based on their intent or attempt to move. Because sensory function is lost or diminished in these individuals, haptic feedback must be non-collocated. The intracortical brain computer interface (iBCI) has enabled a variety of neural prostheses for people with paralysis. An important attribute of the iBCI is that its input signal contains signal-independent noise. To understand the effects of signal-independent noise on a system with non-collocated haptic feedback and inform iBCI-based prostheses control strategies, we conducted an experiment with a conventional haptic interface as a proxy for the iBCI. Able-bodied users were tasked with locating an indentation within a virtual environment using input from their right hand. Non-collocated haptic feedback of the interaction forces in the virtual environment was augmented with noise of three different magnitudes and simultaneously rendered on users' left hands. We found increases in distance error of the guess of the indentation location, mean time per trial, mean peak absolute displacement and speed of tool movements during localization for the highest noise level compared to the other two levels. The findings suggest that users have a threshold of disturbance rejection and that they attempt to increase their signal-to-noise ratio through their exploratory actions. 

%There was no effect of increasing noise magnitude on the frequency properties of movements

\end{abstract}

%%%%%%%%%%%%%%%%%%%%%%%%%%%%%%%%%%%%%%%%%%%%%%%%%%%%%%%%%%%%%%%%%%%%%%%%%%%%%%%%
\section{INTRODUCTION}
Non-collocated haptic feedback has the potential to provide an additional modality of feedback for closed-loop control of motor and communication prostheses. While surface electromyography (sEMG) has been the most common interface method, brain computer interfaces (BCIs) have, more recently, enabled the control of prostheses via decoding neural activity related to the attempt to move. These systems range from non-invasive electroencephalography-based (EEG-based) BCIs to invasive intracortical BCIs (iBCIs). Among the different types of BCIs, the iBCIs have enabled the control of motor prostheses, such as robotic arms \cite{hochberg2012BCIreach} and exoskeletons \cite{ajiboye2017bci}, and communication prostheses \cite{pandarinath2017bci} in both clinical and experimental studies with best in class information throughput. 

  There is an impetus to investigate the effect of non-collocated haptic feedback on iBCI control tasks given its promising results in sEMGs \cite{brown2013understanding}. However, haptic feedback for closed-loop iBCI control operates under unique conditions of signal-independent noisy input signals \cite{willett2017signal} and non-collocation of haptic feedback. The way that these unique conditions might affect user movement strategy is currently unknown; and to date, there has been little exploration of the influence of signal-independent noise on target-acquisition task performance in a system with non-collocated haptic feedback. 

Here we tested the effects of signal-independent noise on the performance of a virtual exploration task with non-collocated haptic feedback. We hypothesize that users are able to reject some level of signal-independent noise in non-collocated haptic feedback while attempting a specified task. Our second hypothesis is that users will attempt to exploit the signal-independency of the noise by increasing the signal-to-noise ratio (SNR) of the feedback through the input-side movements they make.

\begin{figure*}[!t]
\vspace{0.4em}
%\framebox{\parbox{\textwidth}{}}
\centering{\includegraphics[width=0.9\textwidth]{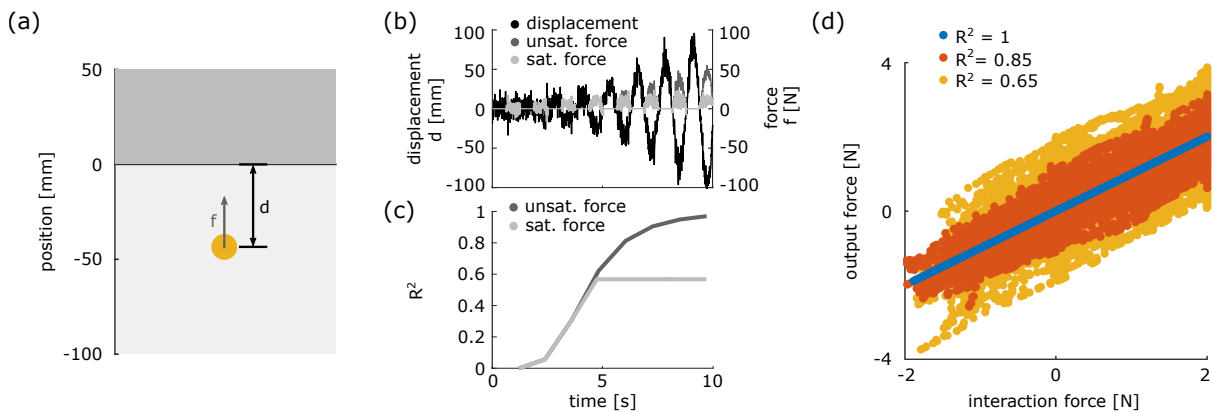}}
%\vspace{-4em}
\caption{(a) Diagram of a simulated user-controlled probe (yellow) coming into contact with an elastic wall (light grey), f (dark grey) is the reaction force due to displacement, d (black). (b) Simulation of a one-dimensional sinusoidal movement of the user-controlled probe with signal-independent noise (black) and the unsaturated noisy resultant force (dark grey) from contacting the elastic wall located at $d = 0$. The light grey signal indicates a scenario where the the noisy resultant force is saturated. (c) Coefficient of determination ($R^2$) between the noisy unsaturated resultant force and the clean resultant force signal (dark grey). $R^2$ increases with displacement magnitude. However if the resultant force is saturated, $R^2$ also saturates (light grey). (d) Scatter plot of output force with signal-independent noise vs. interaction force from one user at different levels of $R^2$ values.}
\vspace{-1em}
\label{noise}
\end{figure*}
% put a line at the elastic wall surface and remove white space

\section{BACKGROUND}
\subsection{Non-collocated Haptic Feedback}

% Section on Non-collocated haptic feedback ------------------------------------------------------------------------
Non-collocated haptic feedback is accomplished by rendering haptic stimulation to a part of the body that is different from the limb that is generating the control signal. In motor and communication prostheses, input signals are based on the users' intention or attempt to move as they have either undergone amputation or lost sensory and motor functions of their limbs due to paralysis. This operating condition results in a unique form of sensorimotor control in which haptic feedback normally accompanying movement is lost or diminished. Thus, prosthetic systems which aim to incorporate haptic feedback as an additional modality must oftentimes provide it on locations of the body where sensitivity is still intact. 

To date, most prior work in non-collocated haptic feedback has focused on sEMG-controlled prostheses that rely on vibrotactile \cite{dalonzo2011vibrotactile}\cite{chatterjee2008vibroprostheses}, skin shear \cite{wheeler2010rotskinstretch}\cite{Kim2012}, or kinesthetic force feedback \cite{panarese2009toesfeedback}\cite{gillespie2010learning_upperlimb} with promising results in grip force modulation, target acquisition, force discrimination and stiffness discrimination tasks. 

For EEG-based BCI, task-related vibrotactile feedback has been shown to improve performance in a virtual cursor control task when visual systems are overloaded \cite{cincotti2007vibrotactile}. Additionally, skin stretch for sensory substitution in EEG-based BCI can improve performance in a cursor targeting task \cite{sketch2015bciskinshear}. In contrast, there has been limited work on non-collocated haptic feedback in iBCIs.

% Section on signal independence --------------------------------------------------------------------------------
\subsection{Signal Independent Noise}

Systems for closed-loop control of iBCIs operate under unique input-output conditions. One such condition is that their output command signals typically contain decoder noise from neural measurements \cite{taylor2002direct}. This decoder noise is signal independent and contributes to the violation of Fitts' Law when tasks are performed over a large enough range of target radii and command signal gains \cite{willett2017signal}. This is opposed to the signal-dependent noise commonly associated with physical movements \cite{meyer1988signaldependentnoise}\cite{guigon2008signaldependentnoise} or sEMG measurements \cite{osu2004signaldepedentEMG}.

In a closed-loop iBCI system, the introduction of signal-independent noise in the command signal results in feedback that also contains signal-independent noise. An analysis of the signal characteristics of iBCI command signals found that the magnitude of noise in the decoded signal had a strong effect on user performance in a targeting task with visual feedback \cite{marathe2015impact}. A non-BCI related kinesthetic force discrimination study found that noise alters a user's perception as measured by Weber fraction \cite{Gurari2017}. 

Signal-independent command signal noise is typically smoothed using Kalman or linear filters \cite{willett2017signal}\cite{marathe2015impact}. However, the introduction of filtering can result in smoothing delays that have been shown to negatively affect performance \cite{marathe2015impact}. In haptic feedback devices for hand-teleoperated systems, delay has been shown to cause a decreased perception of object stiffness though it had no significant effect on grip force modulation \cite{leib2015effect}. In a dynamic reciprocal tapping task, the effect of time delay on haptic feedback resulted in reduced performance. However, the effect is only pronounced at relatively long delay times compared to that of visual delay  \cite{jay2005delayed}. 

\begin{figure}[!b]   
\centering
%\framebox{\parbox{3in}{\centering}}
\vspace{-1em}
\includegraphics[width=0.75\columnwidth]{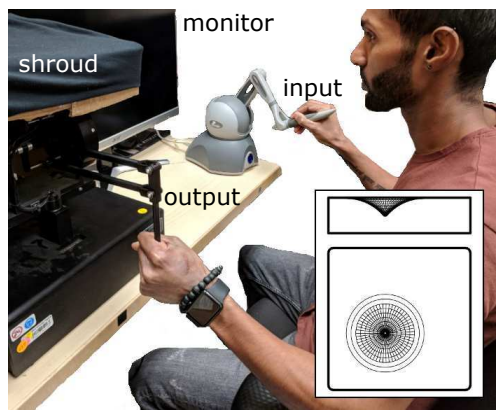}
\caption{Experimental setup. The user holds the input device in the right hand and the non-collocated output device in the left hand. Inset shows an example indentation geometry in the virtual surface (side and top views). }
\label{setup}
\end{figure}

\begin{figure*}[!t]
\vspace{0.4em}
%\framebox{\parbox{\textwidth}{}}
\centering{\includegraphics[width=0.9\textwidth]{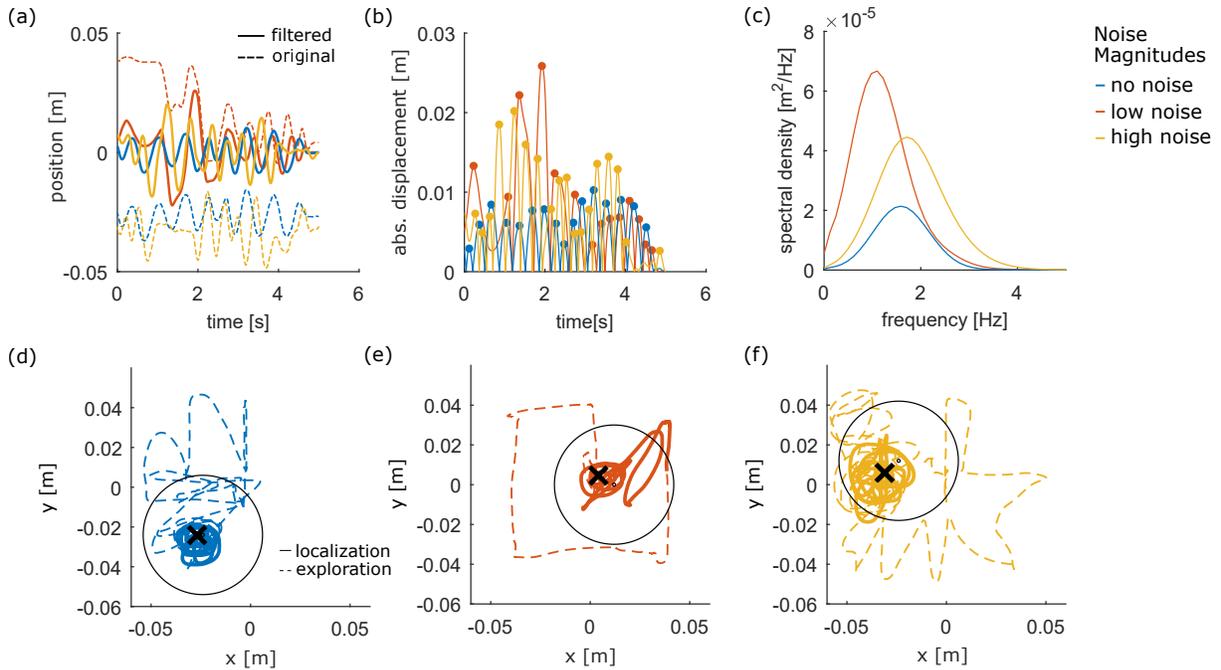}}
\caption{(a) Sample position signals from one subject. (b) Sample absolute position signal after bandpass filtering. Dots indicate peaks that were greater than 10\% of the max peak in that trial. (c) Sample spectral density plots of filtered raw position signals. (d)-(f) Trajectory plots for $R^2=1$ (no noise), $R^2=0.85$ (low noise) and $R^2=0.65$ (high noise). Solid trajectories indicate the last 5 seconds of a trial, when subjects are completing the localization.}
\vspace{-1em}
\label{metrics}
\end{figure*}

\subsection{Signal-to-Noise Ratio Movement Adaptation}

In this work, we predict that users will increase their SNR in response to high signal-independent noise. This form of adaptation is possible because the noise in the system is signal-independent. Hence, increasing the magnitude of user inputs will not result in additional noise. The effect of this strategy is illustrated in Fig.\;\ref{noise}(a) and (b), which shows a simulated sinusoidal user movement of increasing amplitude with constant signal-independent noise periodically coming into contact with an elastic wall at $d=0$. The haptic feedback due to user-generated movements at small amplitudes is indistinguishable from noise but becomes more salient as the amplitude of movement grows. The noise magnitude relative to the input signal can be quantified by the coefficient of determination ($R^2$), which measures the goodness-of-fit of the noisy signal to the clean signal \cite{marathe2015impact}. A lower $R^2$ corresponds to a larger spread of disturbance magnitudes in the signal as shown in Fig.\;\ref{noise}(d). For a fixed standard deviation of Gaussian white noise, the $R^2$ value increases as the maximum user-generated force magnitude increases. Thus in Fig.\;\ref{noise}(c) the $R^2$ value of the resultant force signal approaches its limit of 1 as the amplitude of the displacement signal increases.

\section{METHODS}

\subsection{Hardware and Virtual Environment}

%BCI outputs are decoded from movement-related neural activity. 
As a proxy for an iBCI system, we asked able-bodied users to provide a 3D position input via a Phantom Omni (3D Systems, South Carolina, USA) using their right hand. This input controlled a probe in an environment containing a frictionless virtual tissue sample with a curved indentation rendered by CHAI 3D (Fig.\;\ref{setup}). In the physical environment, this sample had a length of 0.12\;m and a width of 0.12\;m. The indentation had an outer radius of 0.03 m with a depth of 0.0075 m. The user is presented with a top-down view of the environment. The coordinate systems of the input and output devices were aligned with that of the virtual environment. Interaction forces between the probe and the environment were calculated based on the virtual proxy method described in \cite{ruspini1997haptic}. The interaction forces were then injected with noise rendered in real time such that 

\begin{equation}
F_\text{output} = F_\text{interaction} + F_\text{noise}
\label{force_eq}
\vspace{1em}
\end{equation}
with resultant force $F_{\text{output}}$ communicated through a separate Phantom Premium haptic device (3D Systems, South Carolina, USA) that is held in users' left hands (Fig.\;\ref{setup}).

To reduce the confusion between the input and output hands, the users were instructed to hold the input stylus in a pen-like manner and grip the output stylus by wrapping all of their fingers around its cylindrical shape and clasping their thumb over their fingers. To make a guess of the indentation location, users pressed a button located on the stylus of the input device using their index finger. Users were also instructed to attempt to keep their left hand, where haptic feedback was displayed, in the middle of the workspace of the device. 

To prevent the output from saturating, the proxy-goal distance in the virtual environment was limited such that the maximum force that virtual interactions and processed noise signals can produce are each 2 N. 
%The same limit is imposed for the processed noise signals. 
The combined outputs cannot saturate the Premium, which has a limit of 6 N. To isolate the effects of visual feedback, the location of the indentation is occluded during the experiment. Only the projected image of the input device position was shown to aid users' search, without communicating depth of field. 
    
\subsection{Signal-independent Noise Modeling}

The signal-independent noise present in iBCIs can be modeled as Gaussian white noise \cite{willett2017signal}\cite{marathe2015impact}. To provide the flexibility to adjust noise magnitude and the amount of noise signal filtering independently, we adopted the approach in \cite{marathe2015impact}. We generated, in real time, a position disturbance using Gaussian random noise with a mean of 0 and a standard deviation of 1. The signal was then low-pass filtered using a 3rd-order Butterworth filter with a cutoff frequency of 2.5 Hz that was determined through pilot studies. The noise component of the output force was then defined as 
\begin{equation}
 F_\text{noise} = c\, k \, x_\text{noise} 
\label{fxeqn}
\end{equation} 
where $c$ is a scaling constant, $k$ is the virtual material stiffness, and $x_{\text{noise}}$ is the penetration depth due to the position disturbance. We used the coefficient of determination ($R^2$) to quantify the magnitude of noise, as in \cite{marathe2015impact}. The scaling constant $c$ is adjusted such that the noisy force output has a $R^2$ equal to 1 (no noise), 0.85 or 0.65 when compared to the clean output signal with a maximum limited magnitude of 2\;N as in Fig.\;\ref{noise}(d). It is important to note that $R^2$ quantifies the magnitudes of noise over a limited clean output signal range (-2\;N to 2\;N). In contrast, when quantifying SNR using $R^2$, the signal range is not limited but allowed to vary as in the example illustrated in Fig \;\ref{noise}(b).  In this paper, a $R^2$ of 0.85 is considered the ``low noise" condition and a $R^2$ of 0.65 is considered the ``high noise" condition. These $R^2$ values were designed to be in the range of values used in previous iBCI noise simulation experiments \cite{marathe2015impact}. The values of scaling constant $c$ in Eqn.\;\ref{fxeqn} required to achieve the different $R^2$ were determined prior to the experiment.
%\vspace{-0.2cm}

\begin{figure}[!t]
\vspace{0.4em}
\centering
%\framebox{\parbox{3in}{\centering}}
%\vspace{-1em}
\includegraphics[width=0.75\columnwidth]{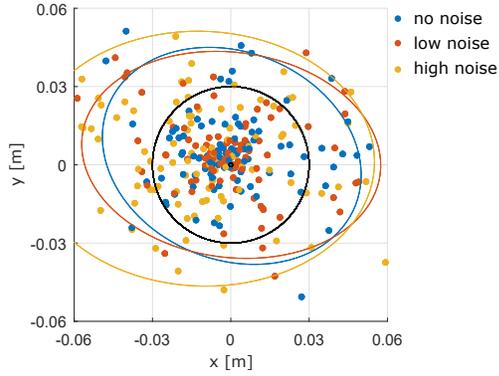}
\caption{Scatter plot of all the guesses made by users with the hole position centered. Ellipses represent the 95\% confidence. Black circle denotes the start of the indentation curvature.}
\vspace{-1em}
\label{scatter}
\end{figure}

\subsection{Experiment Procedures}

Users were asked to identify the position of the center of the indentation using the haptic feedback provided by the system. Prior to beginning the actual trials, users were given 30 seconds of free training time during which their vision of the indentation was not occluded. Next, they proceeded to do 8 mock trials with no noise and no occlusion of the indentation location. Finally, they performed 15 trials with 5 trials at each noise level, with vision occluded. To prevent users from obtaining visual cues from their left hand, the output device and the users' hands were covered by a black shroud for this portion of the experiment. All trials lasted 20 seconds. The order of the different levels of noise was pseudo-randomized in blocks of three. The indentation positions were randomized, with no positions being repeated. The set of indentations used for the actual trials and the training blocks were different. Users were told to prioritize accuracy over speed. They were also instructed to make their best guess before the time ran out. If they failed to move to indicate their best guess using the buttons on the input stylus, their last position at the end of the trial was recorded.

\subsection{Participants}
The experiment was conducted with 20 able-bodied right-handed users between the ages of 22 to 40 who had no existing neuropathies. One other user was excluded from the study because we failed to adequately communicate task instructions due to a language barrier. The protocol for this study was approved by Stanford University's Institutional Review Board and participants gave informed consent.

\subsection{Criteria for Valid Localization Behavior}

To measure the kinematic features of the localization movement, only guesses near the indentation were considered. This is because the indentation was the only region in the virtual space where users were able to feel interaction forces in the xy-plane to perform localization. To be labeled a successful localization within a trial, the administered guess was required to be within the radius (0.03 m) of the indentation. For these valid trials the localization period was then defined to be the last 5 seconds of the trial. All analyses except guess error and time per trial were performed on trials that met this criterion.

\begin{table}[!b]
%\vspace{-1em}
\caption{Table of p-values from Likelihood Ratio Test}
\centering
\begin{tabular}{@{}lcS[table-format=2.3,table-number-alignment = center]S[table-format=1.3,table-text-alignment=center]@{}}
\toprule
Metric                                       & \multicolumn{1}{l}{} & \multicolumn{1}{c}{Noise Magnitude} & \multicolumn{1}{c}{Progression} \\ \midrule
                                             & df                   & \multicolumn{1}{c}{1}                                    & \multicolumn{1}{c}{1}                                \\ \midrule
\multirow{2}{*}{Guess Distance Error}                 & LR                  & 19.62                               & 0.68                            \\
                                             & p                    & \textbf{\hspace{1.8em}\textless{}0.001$^{***}$}           & 0.408                           \\ \midrule
\multirow{2}{*}{Time per Trial}              & LR                  & 17.89                               & 3.67                            \\
                                             & p                    & \textbf{\hspace{1.8em}\textless{}0.001$^{***}$}           & 0.054                           \\ \midrule
\multirow{2}{*}{\parbox{3cm}{Mean Peak Displacement during Localization}}           & LR                  & 6.63                                & 1.44                            \\
                                             & p                    & \textbf{\hspace{2.7em}0.038$^*$}                      & 0.229                        \\ \midrule
\multirow{2}{*}{\parbox{2.5cm}{Mean Peak Speed during Localization}}                  & LR                  & 7.71                                & 1.96                            \\
                                             & p                    & \textbf{\hspace{2.7em}0.022$^*$}                      & 0.167                           \\ \midrule
\multirow{2}{*}{\parbox{3cm}{Mean Peak Output Force during Localization}}           & LR                  & 9.10                                & 7.85                            \\
                                             & p                    & \textbf{\hspace{2.7em}0.011$^*$}                      & \hspace{1.6em}\textbf{0.005$^{**}$}                  \\ \midrule
\multirow{2}{*}{\parbox{2.9cm}{Mean Peak Disp. Freq. during Localization}}           & LR                  & 0.28                                & 1.79                            \\
                                             & p                    & 0.596                      & 0.189                  \\ \midrule
\multirow{2}{*}{\parbox{2.7cm}{Mean Peak Speed Freq. during Localization}}           & LR                  & 4.93                               & 1.79                            \\
                                             & p                    & 0.087                     & 0.31                 \\ \midrule                                             
\multirow{2}{*}{\parbox{3cm}{Mean Output Force Freq. during Localization}} & LR                  & 0.28                                & 1.16                            \\
                                             & p                    & 0.088                               & 0.307                             \\ \bottomrule

\end{tabular}
\label{ANOVAtable}
\end{table}

\begin{figure*}[!t]
\vspace{0.4em}
\centering
%\framebox{\parbox{\textwidth}{}}
\centering\includegraphics[scale=0.8]{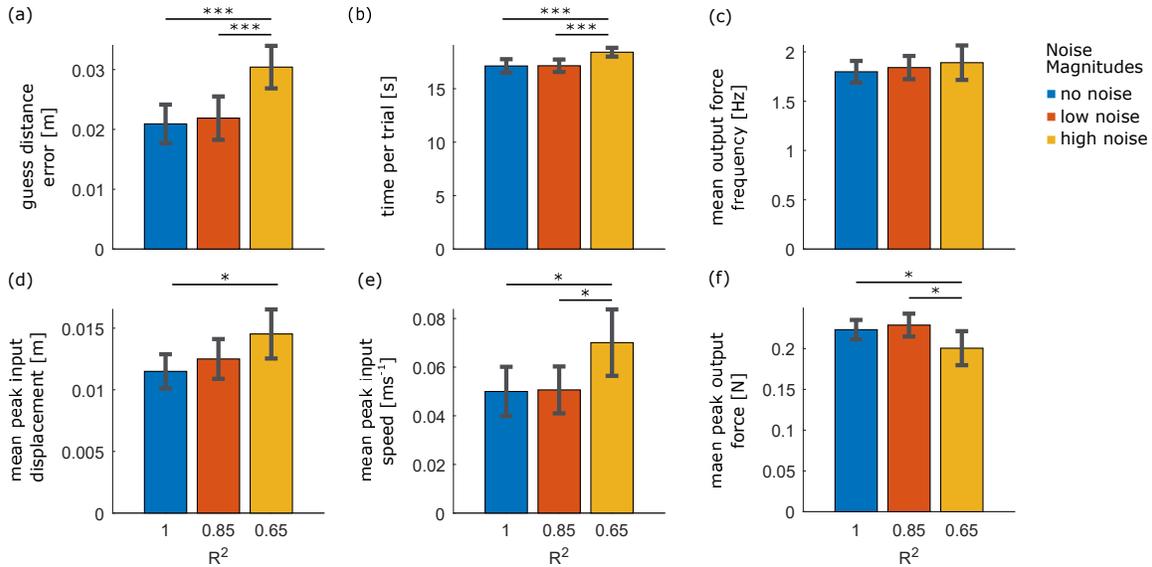}
\caption{Results of selected metrics across different noise magnitudes. (a) Guess distance error. (b) Time elapsed per trial. (c) Mean frequency of output interaction force during localization. (d) Mean absolute peak displacement in dominant movement direction during localization. (e) Mean peak speed during localization. (f) Mean peak output interaction force magnitude during localization. All kinematic units are in the scale of the physical environment. All data is comprised of components in the xy-plane (parallel to surface of the tissue). Error bars denote 95\% C.Is. 
$* : \text{p-value}<0.05 \ ** : \text{p-value}<0.01 \ *** : \text{p-value}<0.001$}
\label{results}
\vspace{-1em}
\end{figure*}

\subsection{Metrics}

For this study, we considered several metrics to analyze the changes in users' movement strategy in response to the varying levels of signal-independent noise. 

To quantify the amount of error of users' guesses with respect to the actual position of the indentation, the distance between the indentation center and each guess in the plane parallel to the surface of the virtual tissue (xy-plane) was measured for each trial.

Peak analysis was used to quantify the kinematic properties of the users' input-side position variations about their moving average in the xy-plane. The absolute valued peak displacements averaged over the localization period of a single trial gives a measure of a user's input signal magnitude for that trial. 

Prior to performing peak analysis, each component of position was filtered using a 3rd-order Butterworth bandpass filter with a passband of 0.5\ Hz to 10\ Hz. This removed the high frequency noise and the DC component of the cursor position data. The absolute values of the signal in each component direction, horizontal and vertical, were taken and the peak amplitudes computed using the MATLAB function \textit{fpeaks.m}. The minimum peak height and prominence were set to 10\% of the maximum peak height of that trial. The variability of users' trajectories resulted in the masking of signal peaks in either component direction by the other if the magnitude of the vector was used for analyses. To eliminate this effect, only the dominant component of users' movements, either horizontal or vertical, was considered. This was determined by calculating the mean absolute distance in each component direction for each user and each trial, with the dominant component being the one with the larger mean. An example of the peak analysis for one subject for three trials of varying noise levels is shown in Figs.\;\ref{metrics}(a) and (b). 

To further understand other changes in users' movement strategies, peak input movement speeds were averaged over the localization period within each trial. In addition, the de-noised peak interaction forces were also averaged over the localization period of each trial to understand the force feedback the user was deriving from the input movements. For both these metrics, the magnitude of the components in the xy-plane of the virtual tissue was filtered using a 3rd-order Butterworth low-pass filter with a cutoff frequency of 10 Hz. The filtered vector magnitudes were then analyzed using peak analysis with minimum peak height set to 10\% of the maximum of that trial and averaged over the localization period.

As a complement to temporal analyses, we also conducted spectral analyses of users' movements. The dominant component of filtered input position, and in addition, the xy-plane input speed and the de-noised output interaction force magnitude during the localization period were also analyzed in the frequency domain using Welch's Power Spectral Density (PSD) Estimate. This was done using the MATLAB function \textit{pwelch.m} with the default Hamming window. To eliminate the DC components of the speed and force signals, their averages were removed using the MATLAB function \textit{detrend.m} prior to performing the PSD estimate. The mean frequency of the spectral density distribution was computed using the MATLAB function \textit{meanfreq.m}. An example spectral density plot for position signals is shown in Fig.\;\ref{metrics}(c).

\subsection{Statistical Analysis}

A linear mixed effects model was fit to each response variable. The fixed effects were the noise magnitude and the experiment progression. The random effect was the subject fitted as an intercept. The model did not fit for interaction between the two fixed effects because it was not found to be significant over all response variables. A likelihood ratio test with a parametric bootstrap of 10000 simulations was conducted to test for significance of the fixed effects. If the effect of noise magnitude was significant with $p<0.05$, a post-hoc test with Bonferroni correction was performed to evaluate the significance of pairwise differences between levels.

%If a significance level of $p < 0.05$ was found for the effect of noise level, a post-hoc test with Bonferroni correction was performed to evaluate the pairwise differences between levels.

\vspace{0.5em}

\section{RESULTS}

The likelihood ratio test found the overall effect of noise magnitude on mean distance error and mean time per trial to be significant. 

As shown in Fig.\;\ref{results}(a), the mean distance error of user guesses was higher for the condition with the highest noise ($R^2=0.65$) than compared with the error for the low noise and no noise conditions ($R^2 = 0.85 \textrm{ and } 1$ respectively). Post-hoc tests showed that the effect was significant for both comparisons ($p<0.001$). A graphical representation of the spread of the guesses is shown in Fig.\;\ref{scatter}. 

The amount of time taken per trial was also higher for the high noise condition compared to the no noise and low noise condition (Fig.\;\ref{results}(b)), with the effect being significant for both comparisons (${p<0.001}$). 

The application of the criterion for valid localization behavior yielded 77 trials for the no noise, 74 for the low noise and 52 for the high noise condition.

The mean peak displacement during localization increased as noise magnitude increased (Fig.\;\ref{results}(d)). This effect was found to be significant. Post-hoc tests showed that the no noise condition had significantly higher mean peak displacement than the high noise condition ($p=0.032$). Noise magnitude level was shown to have a significant effect on mean peak speed during localization (Fig.\;\ref{results}(e)). The post-hoc tests showed there was significantly higher mean peak speed in the high noise condition than in the no noise ($p=0.035$) and low noise condition ($p=0.032$). Mean peak output interaction force during localization was also significantly affected by noise magnitude (Fig.\;\ref{results}(f)), with a significantly lower mean peak output force in the high noise condition than in the no noise ($p=0.020$) and low noise noise ($p=0.015$) conditions. 

None of the mean frequency metrics from the spectral density analysis showed significant differences between the noise magnitude levels (Fig.\;\ref{results}(c)).

For all metrics, the experiment progression did not show any significant effects during localization across the noise magnitude levels except for that of the mean peak output force. The results of the likelihood ratio test are summarized in the form of the degrees of freedom (df), Likelihood Ratio test statistic (LR) and p-values (p) in Table \ref{ANOVAtable}.

\section{DISCUSSION}

Our results suggest that, as hypothesized, users of non-collocated haptic devices are able to reject signal-independent noise up to a threshold magnitude. The significant increase in guess error and time per trial for the high noise condition over the other two conditions indicate that the effect of noise on task performance is significant when $R^2$ decreases from 0.85 to 0.65. Therefore, in this particular task, the SNR threshold, as quantified by $R^2$, falls between 0.65 and 0.85. It is important to note that it is possible that the SNR at which task performance degrades significantly found here is task specific. 
%and thus future studies should include exploring the thresholds over more levels and for other tasks such as center-out reaching.

Qualitatively, the low SNR manifests as high uncertainty. During the post-experiment survey users reported being unsure about the feedback they were getting. They attributed this to their difficulty in consistently separating the interaction forces from the noise.

Our results also support our second hypothesis that users attempt to increase the SNR of the output to increase the saliency of the interaction forces. While performing the task under high noise magnitude, users modified their localization strategy by increasing the amplitude and speed of their movements (Figs.\;\ref{results}(d) and (e)). The simulation in Fig.\;\ref{noise}(a)-(c) shows that this increase in movement amplitude increases the SNR as measured by $R^2$.

  Despite users' best efforts, their attempts to boost SNR were ineffective due to force rendering limitations. While there was a statistically significant decrease in the mean peak output interaction force magnitude (Fig.\;\ref{results}(f)), when all three components of force are considered as a scalar magnitude, the difference is eliminated. The lack of meaningful variation in the peak interaction forces despite larger input movements points towards the users reaching the imposed 4\;N total force rendering limit. With this saturation, users cannot increase the SNR as measured by $R^2$ by increasing the magnitude of their input movements. The saturated noisy resultant force and its corresponding $R^2$, shown as light grey in Fig.\;\ref{noise}(b) and (c), illustrate this concept in simulation.
  
 The increases in mean peak speed during localization suggests that the mean frequency characteristics of the output force would be higher. However, this was not true: All mean frequency metrics calculated showed no significant difference across noise magnitudes. This may be due to low frequency components with high magnitudes causing the mean frequency to be less sensitive to variation in higher frequency components. Since this experiment did not vary noise frequency, users might not have focused on varying their frequency as much as attempting to vary their force amplitude.
 
\section{CONCLUSIONS AND FUTURE WORK}

In this work, we measured the effects of adding signal-independent noise to non-collocated haptic feedback on the performance of able-bodied users doing a virtual exploration and feature localization task. Our experimental setup measured input movements from the right hand and rendered virtual interaction forces with predefined magnitudes of signal-independent noise on the left hand.

Results showed that users could achieve similar levels of performance provided the signal-to-noise ratio was below a certain threshold as measured by $R^2$ values. Designers of future robotic prostheses with non-collocated haptic feedback and signal-independent noise, such as those using iBCI, can possibly tune the amount of signal smoothing to achieve higher system bandwidth at the expense of some low level noise in the feedback loop.

In addition, we interpret the coefficient of determination ($R^2$), often used in iBCI literature to quantify relative noise magnitude, as a measure of signal-to-noise ratio. We found that users will attempt to increase their SNR when using systems with high noise. Thus if iBCI haptic systems are designed with adequate margins of force display, then users can leverage the unique signal-independent nature of the system noise to improve the saliency of their feedback by changing their movements.

%\begin{figure}[!t]   
%\centering
%%\framebox{\parbox{3in}{\centering}}
%%\vspace{-1em}
%\includegraphics[width=0.75\columnwidth]{conclusion_plot.eps}
%\caption{(a) Simulation of a one-dimensional sinusoidal movement of the user-controlled probe with signal-independent noise (blue) and the unsaturated resultant force (red) and saturated resultant force (yellow) from contacting the elastic wall located at $d=0$. (b) Coefficient of determination ($R^2$) between the noisy resultant force and the clean resultant force signal with (yellow) and without (red) saturation. }
%\vspace{-1em}
%\label{SNR_conclusion}
%\end{figure}

In this study, noise frequency was not varied. This may explain why the frequency of users' movements and force outputs did not vary significantly. Future work should investigate the effect of varying frequencies of signal-independent noise. 

Users of robotic prostheses rely heavily on visual feedback alone when performing control tasks. This study shows that users can leverage haptic sensory pathways to interpret task-relevant information even when confronted with noise. Moving forward, haptic feedback will be an important modality to consider in the design of iBCI-based prostheses systems in order to achieve naturalistic control and user system embodiment.

\addtolength{\textheight}{-9.5  cm}   % This command serves to balance the column lengths

%%%%%%%%%%%%%%%%%%%%%%%%%%%%%%%%%%%%%%%%%%%%%%%%%%%%%%%%%%%%%%%%%%%%%%%%%%%%%%%%

%%%%%%%%%%%%%%%%%%%%%%%%%%%%%%%%%%%%%%%%%%%%%%%%%%%%%%%%%%%%%%%%%%%%%%%%%%%%%%%%

%%%%%%%%%%%%%%%%%%%%%%%%%%%%%%%%%%%%%%%%%%%%%%%%%%%%%%%%%%%%%%%%%%%%%%%%%%%%%%%%

%%%%%%%%%%%%%%%%%%%%%%%%%%%%%%%%%%%%%%%%%%%%%%%%%%%%%%%%%%%%%%%%%%%%%%%%%%%%%%%%

%\begin{thebibliography}
\bibliographystyle{IEEEtran}
\bibliography{IEEEabrv,biblio}
%\end{thebibliography}

\end{document}